\PassOptionsToPackage{table}{xcolor}
\documentclass{article}
\usepackage{iclr2027_conference,times}

\usepackage[english]{babel}

\usepackage{amsmath}
\usepackage{amssymb}
\usepackage{array}
\usepackage{pifont}
\usepackage{lipsum}
\usepackage{graphicx}
\usepackage{xcolor}
\usepackage[colorlinks=true, allcolors=blue]{hyperref}
\usepackage{doi}
\usepackage{booktabs}
\usepackage{tikz}
\usetikzlibrary{positioning}
\usepackage{multirow}

\definecolor{rankgold}{HTML}{F7E3A1}
\definecolor{ranksilver}{HTML}{E4E4E4}
\definecolor{rankbronze}{HTML}{F3DCC8}
\newcommand{\first}[1]{\cellcolor{rankgold}\textbf{#1}}
\newcommand{\second}[1]{\cellcolor{ranksilver}\underline{#1}}
\newcommand{\third}[1]{\cellcolor{rankbronze}#1\textsuperscript{\dag}}
\newcommand{\swatch}[2]{{\setlength{\fboxsep}{2pt}\colorbox{#1}{#2}}}
\newcommand{\keyfirst}{\swatch{rankgold}{\textbf{bold gold}}}
\newcommand{\keysecond}{\swatch{ranksilver}{\underline{underline silver}}}
\newcommand{\keythird}{\swatch{rankbronze}{bronze\dag}}  
\newcommand{\gn}[1]{\textcolor{green!55!black}{#1}}
\newcommand{\rd}[1]{\textcolor{red!75!black}{#1}}
\newcommand{\gy}[1]{\textcolor{black!40}{#1}}

\title{No Distillation Needed: Single-Pass Real-Time Talking Heads via Acausal Noise Shaping}

\author{{\bf Yu Han, Dejan Markovic, Alexander Richard, Wojciech Zielonka,} \\
{\bf Akshay Venkatesh, Cheng-hsin Wuu \& Michael Zollhoefer} \\
Meta Reality Labs \\[2mm]
\centerline{\href{https://wojciechzielonka.com/facegan/}{Project Website}}
}

\iclrfinalcopy 
\begin{document}
\maketitle
\lhead{Preprint.} 

\begin{figure}[h]
    \vspace{-1.1cm}
    \centerline{\includegraphics[width=1.0\textwidth]{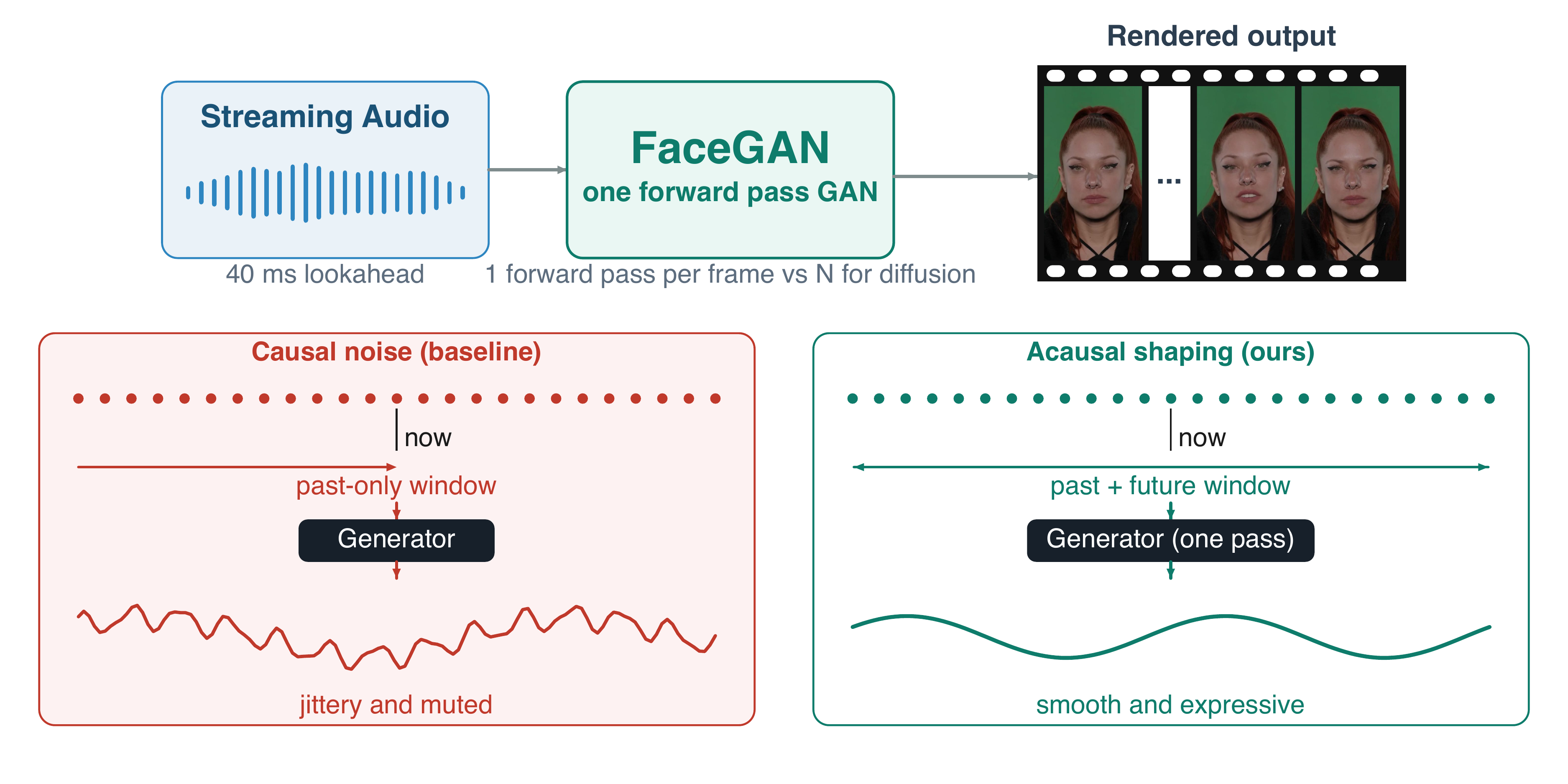}}
    \vspace{-0.75cm}
    \caption{Streaming audio-to-expression with latency-free acausal noise shaping. Top: audio streams through our method called \textbf{FaceGAN}, which emits every frame in a single forward pass at 25 Hz with 40 ms of audio lookahead; the filmstrip shows rendered output. Bottom: the same causal generator driven by causal noise (left) produces jittery, muted motion, while our acausally shaped noise (right) yields smooth, expressive motion. Bottom traces are illustrative schematics.}
    \label{fig:teaser}
\end{figure}

\begin{abstract}
Audio-driven facial animation underpins real-time avatars, telepresence, and embodied virtual agents. And it must run online: each frame emitted from audio observed up to the current time, at interactive rates. Recent progress is dominated by diffusion models, which need many network evaluations per sample and are therefore a poor fit for streaming. We argue the cost is unnecessary in this domain. Audio-conditioned facial motion occupies a comparatively low-dimensional manifold, a regime where a single-pass GAN suffices. The obstacle is not capacity but stochastic structure. We show that a causal, time-invariant generator driven by i.i.d.\ noise cannot suppress its output spectrum over a band without collapsing its per-step innovation. We proposed \textbf{FaceGAN}, which dissolved the limitation by shaping the noise pathway acausally. Because the driving noise is synthetic, its future can be sampled now, so the audio-to-expression path stays causal, and the model supports fully causal operation. \textbf{FaceGAN} emits expression and head pose in a single forward pass per frame and matches or outperforms state-of-art approaches in generation quality. Being feed-forward with bounded attention windows, it generates indefinitely without drift.
\end{abstract}

\section{Introduction}

Audio-driven facial expression generation, mapping a stream of speech to a stream of facial motion, is a core building block for virtual avatars, telepresence, embodied agents, game characters, and assistive communication. In practical settings the model runs online: it must emit expression frame $t$ from audio observed up to that time and small lookahead, at interactive frame rates, with bounded and small latency. Offline fidelity is not enough; the deployment constraint is streaming inference.

The task itself is mature, and recent gains have been driven largely by diffusion models, which produce high-fidelity, diverse motion but at the cost of iterative sampling: tens to hundreds of function evaluations per frame. This is a poor fit for streaming: the very mechanism that gives diffusion its quality is what makes it expensive to deploy online.
The prevailing response is to accept the cost, or to distill diffusion back down toward a single pass, implicitly conceding that single-pass generation is the desirable endpoint and diffusion is a detour to reach it.

We take the endpoint directly. Our starting observation is that diffusion's advantages are most pronounced on high-dimensional, highly multimodal data distributions; audio-conditioned facial expression, by contrast, lives on a comparatively low-dimensional manifold and is typically learned under data scarcity. In this regime the gap that motivates iterative refinement largely disappears, and a single-pass generator (concretely, a GAN) can match or exceed diffusion quality while retaining single-pass, low-latency inference. In other words, the efficient, well-understood tool is sufficient here; the question is only whether it can be made to produce natural motion online.

The obstacle to that is not obvious, and it is where our second contribution lies. A naive causal single-pass generator faces an intrinsic spectral limitation: driven by i.i.d. noise, a causal, time-invariant generator cannot shape the spectrum of its output without paying in innovation variance. We make this precise via the theorem in Sec.~\ref{sec:spectral_limitation}: suppressing the output spectrum over a band forces the per-step new-information budget to collapse. Empirically this manifests as the trade-off practitioners hit: either the motion is temporally rich but jittery, or smooth but muted and lifeless. Increasing network capacity does not help -- it is a property of the generated process, not the architecture.

Our fix is a single, cheap architectural move: acausal shaping of the noise pathway. Because the driving noise is generated internally rather than observed from the environment, an acausal filter over the noise can be realized with zero latency: we simply sample the required future noise now. This makes the noise fed to the generator non-i.i.d. The result is a system that is causal and streaming in the audio$\to$expression pathway, yet free to produce smooth, spectrally clean, and richly stochastic motion. The acausal stage is generator-agnostic in principle and may benefit other generators; whether it helps diffusion is orthogonal to our thesis and left to future work, and even if it did, diffusion's iterative sampling cost would remain. 

Together these two threads, the efficiency of single-pass GANs on this manifold and the noise-shaping technique that unlocks their quality, yield FaceGAN, a low-latency, single-pass audio-to-expression system that matches or surpasses diffusion baselines at a fraction of the inference latency.

Our main contributions are as follows:

\begin{itemize} 
\item We identify and formalize an intrinsic spectral limitation to causal single-pass generation via the Kolmogorov--Szeg\H{o} theorem, explaining the jitter-vs-muted-motion trade-off as a property of the generated process rather than the architecture.
\item We introduce latency-free acausal noise shaping, which dissolves this limitation by driving a causal generator with non-i.i.d. noise, and show it costs no audio latency.
\item We build FaceGAN and show that on the low-dimensional manifold of audio-driven facial expression, a single-pass GAN matches the compared baselines at $43$\,ms latency and $12\times$ real time on a single GPU. 
\end{itemize}

\section{Related work}

\subsection{Audio-Driven Facial Motion Generation}

Audio-driven facial motion generation synthesizes facial movement from speech, and has been studied across output representations ranging from 2D portrait videos~\cite{zhou2020makeittalk, prajwal2020lip} to 3D mesh templates~\cite{karras2017audio, fan2022faceformer, richard2021meshtalk, xing2023codetalker, thambiraja2023imitator} and, more recently, neural rendering with NeRF~\cite{guo2021ad, ye2023geneface, gafni2021nerface, zielonka2023insta} or Gaussian Splatting~\cite{aneja2025gaussianspeech, li2024talkinggaussian, zielonka2025gem, zielonka2025synshot, qian2024gaussianavatars} for higher visual fidelity (see \cite{zielonka2026star} for a comprehensive survey of avatar representations and creation methods across face, full-body, hands, hair, and garments). Orthogonal to the choice of representation, a central difficulty is that audio only weakly determines facial dynamics: deterministic regressors collapse toward the conditional mean, producing over-smoothed, under-expressive motion that discards a speaker's stochastic detail. To recover this diversity, recent work turns to generative models. Diffusion approaches~\cite{aneja2024facetalk, sun2024diffposetalk} synthesize high-fidelity, varied motion but depend on full-sequence look-ahead or iterative sampling, and even streaming-oriented~\cite{MotionStreamer_2025_ICCV} or single-step~\cite{lee2025audio} variants operate at chunk granularity. Autoregressive token models predict discrete motion codes step by step~\cite{chu2024artalk}, which~\cite{Fallingwater} pushed to strictly causal, zero-lookahead streaming. Both families restore motion diversity, but each carries an inference cost: diffusion pays for iterative refinement, and autoregressive models require a learned tokenizer and sequential, token-by-token decoding. 

\paragraph{Our setting.} We target streaming, single-pass generation: each frame is emitted from audio observed up to the current time, in one forward pass, at low latency. This rules out full-sequence and chunk-based methods, and distinguishes us from autoregressive streaming approaches~\cite{chu2024artalk, Fallingwater}, which remain sequential in the token dimension. We show that on the comparatively low-dimensional manifold of facial motion, a single-pass GAN matches or exceeds these approaches at a fraction of the inference cost, provided the stochastic degradation that afflicts cheap single-pass generators is addressed, which we do via latency-free acausal noise shaping (Sec.~\ref{sec:tradeoff}).

\subsection{Generative Models for Motion: Diffusion vs. Single-Pass Generator}

The choice of generative family reflects a trade-off between sample quality and inference cost. Diffusion models became the default for high-dimensional, highly multimodal domains, and continue to scale through flow-based formulations~\cite{lipman2023flow, esser2024scaling, flux2024}, where iterative denoising both stabilizes training and captures complex, multi-peaked distributions. This power comes at a price: sampling requires many sequential network evaluations. Consequently, a large and very active line of work seeks to compress diffusion back toward a single step: consistency models~\cite{song2023consistency, luo2023latent} and their continuous-time refinement~\cite{lu2025simplifying}, trajectory- and shortcut-based samplers~\cite{kim2024ctm, frans2025shortcut, geng2025meanflow}, and distillation objectives such as distribution matching~\cite{yin2024dmd, yin2024dmd2} and adversarial distillation~\cite{sauer2023adversarial, xu2024ufogen}. That so much recent effort targets one-step sampling underscores the point: single-pass generation is the desirable inference regime, and these methods are routes back to it. Notably, several of these state-of-the-art accelerators converge on ideas from GANs. Adversarial objectives are reintroduced to sharpen few-step samples~\cite{sauer2023adversarial, xu2024ufogen}, and recent work distills diffusion models directly into conditional GANs~\cite{kang2024distilling}, while modernized GAN architectures and training recipes~\cite{kang2023scaling, sauer2023stylegant, huang2024gan} have narrowed or closed the quality gap with diffusion at a single forward pass. GANs remain particularly strong on lower-dimensional or strongly structured domains and under limited data~\cite{karras2020training}, where diffusion's multi-step machinery offers diminishing returns and can even overfit. Face motion is such a regime: conditioned on audio, expression and head pose occupy a comparatively low-dimensional manifold, and paired audio-motion data is scarce relative to image or video corpora, so a single-pass conditional generator can match iterative samplers in quality while retaining a latency advantage for streaming synthesis.

\begin{table}[!t]
  \centering
  \small
  \vspace{-0.5cm}
  \begin{tabular}{l|c|c|c|c}
  Method & Arch. & \# steps & Causal & \shortstack{Live-system} \\\hline
  DiffPoseTalk~\cite{sun2024diffposetalk} & Diffusion & 500 & \textcolor{orange}{$\sim$} (chunk-level) & \textcolor{red}{\ding{55}} \\
  ARTalk~\cite{chu2024artalk} & LLM & 5 & \textcolor{orange}{$\sim$} (chunk-level) & \textcolor{red}{\ding{55}} \\
  MemoryTalker~\cite{MemoryTalker} & Transformer & 1 & \textcolor{red}{\ding{55}} & \textcolor{red}{\ding{55}} \\
  Fallingwater~\cite{Fallingwater} & LLM & 176 & \textcolor{orange}{$\sim$} (chunk-level) & \textcolor{red}{\ding{55}} \\
  \textbf{FaceGAN (ours)} & \textbf{GAN} & \textbf{1} & \textbf{\textcolor{green!55!black}{\ding{51}}} (frame-level) & \textbf{\textcolor{green!55!black}{\ding{51}}}
  \end{tabular}
  \caption{\label{tab:inference}Inference characteristics of the compared methods. Only our method is both single-pass and frame-level streaming; see Appendix~\ref{app:inference-steps} for how \# steps is counted. Live-system means deployable for live animation: frame-level streaming with low latency, not throughput alone; see Appendix~\ref{app:realtime-benchmark} for the measured benchmark.}
  \vspace{-10pt} 
\end{table}

\paragraph{Our position.} We therefore build on a single-pass GAN rather than an iterative or autoregressive sampler. The obstacle is not quality but stochastic structure: a causal, single-pass generator driven by i.i.d. noise faces an intrinsic limit on how it can shape the temporal spectrum of its output without sacrificing motion richness (Sec.~\ref{sec:tradeoff}). We show this limitation, not any inherent weakness of GANs, is what has kept cheap single-pass generators uncompetitive for temporally coherent motion, and that it is removed by shaping the noise pathway acausally at no cost to audio-side latency. Table~\ref{tab:inference} summarizes how the compared methods differ along these axes.

\section{Spectral Limitation of Causal Generation}
\label{sec:spectral_limitation}
A \textbf{causal} and \textbf{time-invariant} generator maps i.i.d.\ noise $z=(z_t)_{t\in\mathbb{Z}}$ to outputs $y=(y_t)_{t\in\mathbb{Z}}$ via
\begin{equation}
  y_t = F(z_t, z_{t-1}, z_{t-2}, \ldots),
  \label{eq:generator}
\end{equation}
where $F$ is any measurable function (an arbitrary neural network). For natural facial motion $y$ should have \textbf{bandlimited} spectrum: energy concentrated at low frequencies, and none above a cutoff. The question is whether such a causal $F$ can produce a perfect stopband (exact spectral null) in $y$. The linear case is settled by Paley--Wiener theory~\cite{paley1934fourier}; and here we show a nonlinear $F$ cannot do it either: the argument below applies to the generated \emph{stochastic process} rather than to any particular architecture, and therefore binds arbitrary causal $F$ without linearization. The argument proceeds in two parts. We first prove impossibility of a perfect stopband via Szeg\H{o}'s link between spectral nulls and predictability, then quantify the depth-vs.-innovation trade-off and show how acausal noise shaping escapes it. Background is recalled in Appendix~\ref{app:prereq}; we use \eqref{eq:nondeg}--\eqref{eq:totalvar} below.

\subsection{No perfect spectral stopband}

\textbf{Theorem (Kolmogorov--Szeg\H{o}).}~\cite{szego1920beitrage,kolmogorov1941stationary,pourahmadi2001foundations} Let $y_t$ be a real $L^2$-stationary stochastic process with spectral density $S : [-\pi, \pi] \to [0,\infty)$. Then the asymptotic one-step linear prediction error variance equals
\begin{equation}
\sigma_\infty^2 = \exp\!\left(\frac{1}{2\pi}\int_{-\pi}^{\pi} \log S(\omega)\,d\omega\right), \quad \exp(-\infty):=0.
\label{eq:KolmogorovSzego}
\end{equation}

\textbf{Key consequence.} If $S=0$ on a set $E\subseteq [-\pi, \pi]$ of positive measure, then $\log S=-\infty$ on $E$, so $\int\log S\,d\omega=-\infty$ and $\sigma_\infty^2=0$ by \eqref{eq:KolmogorovSzego}. The process is then deterministic~\cite{doob1953stochastic}, so $\mathrm{Var}(y_t\mid y_{<t})=0$.

\textbf{Theorem (main claim).} Assume the generator uses its current noise, i.e.\ $\mathrm{Var}(F(z_t,z_{<t})\mid z_{<t})>0$ a.s.\ (cf.\ \eqref{eq:nondeg}). Let $y$ be the output in \eqref{eq:generator}. Then its spectral density satisfies $S_y(\omega)>0$ for a.e.\ $\omega$. In particular, $S_y$ has no perfect stopband.

\textbf{Proof of main claim.} By \eqref{eq:totalvar}, $\mathrm{Var}(y_t\mid y_{<t})\geq\mathbb{E}[\mathrm{Var}(y_t\mid z_{<t})\mid y_{<t}]$. Given $z_{<t}$, $y_t$ depends only on $z_t$, so $\mathrm{Var}(y_t\mid z_{<t})>0$ a.s.\ by \eqref{eq:nondeg}; conditioning on $y_{<t}$ preserves strict positivity, so $\mathrm{Var}(y_t\mid y_{<t})>0$ a.s. Thus $y_t$ lies outside the closed linear span of $y_{<t}$ and $\sigma_\infty^2>0$, which by the key consequence rules out any positive-measure null of $S_y$. \hfill$\square$

\subsection{Quantitative trade-off: stopband depth vs.\ innovation}
\label{sec:tradeoff}

Suppose $S_y\leq\varepsilon$ over a fraction $\delta\in(0,1)$ of the spectrum, with $S_y$ scaled so its peak is $1$ (hence $\log S_y\leq 0$ elsewhere). Then from \eqref{eq:KolmogorovSzego},
\begin{equation}
\log\sigma_\infty^2=\frac{1}{2\pi}\int_{-\pi}^{\pi}\log S_y(\omega)\,d\omega\leq\delta\cdot\log\varepsilon
\quad\Longrightarrow\quad
\boxed{\sigma_\infty^2\leq\varepsilon^{\delta}.}
\label{eq:quant}
\end{equation}

For example, for a 40\,dB stopband ($\varepsilon=10^{-4}$ in power) over $\delta=0.9$ of the spectrum, $\sigma_\infty^2\leq10^{-3.6}\approx 2.5\times 10^{-4}$: at most $\sim$0.025\% of input noise variance survives as new per-step information. Insisting on rich stochasticity (large $\sigma_\infty^2$) forces a leaky stopband (jitter); insisting on small $\varepsilon$ collapses $\sigma_\infty^2$ (muted motion); buying back both with deep causal memory costs latency.

\paragraph{Escaping via acausal noise shaping.} The limitation binds i.i.d.\ driving noise via $\mathrm{Var}(F(z_t,z_{<t})\mid z_{<t})>0$ a.s.\ (cf.\ \eqref{eq:nondeg}). Preprocess acausally $z\mapsto\hat{z}$: $\hat{z}_t$ becomes predictable from $\hat{z}_{<t}$ since the smoother mixes in future $z$, so \eqref{eq:nondeg} fails for $\hat{z}$ by design and the downstream causal $F$ in \eqref{eq:generator} is no longer bound by \eqref{eq:quant}. Since $z$ is synthetic, sampling its future needs no audio lookahead.

\section{Method}
\begin{figure}[h]
    \centering
    \includegraphics[width=\textwidth]{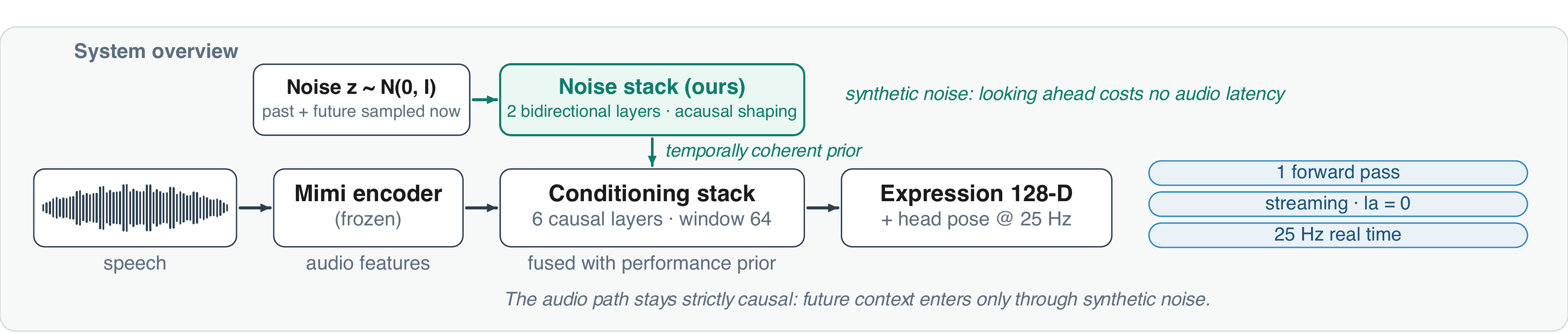}
    \caption{System overview. Speech is encoded by a frozen Mimi encoder while an independent noise pathway samples past and future noise (costing no latency, as the noise is drawn rather than observed) and shapes it with two bidirectional layers into a temporally coherent motion prior. A six-layer causal conditioning stack fuses the audio features with the prior and emits 128-D expression codes plus head pose at 25 Hz in a single forward pass, with no iterative sampling.}
    \label{fig:system}
    \vspace{-10pt}
\end{figure}

\textbf{Motion representation.} We operate in the same motion latent space as \cite{seamless2025}, using their pretrained encoder. Each frame $i$ is encoded into an expression code $\mathbf{e}_i \in \mathbb{R}^{128}$, head rotation $\mathbf{r}^{\text{h}}_i$, head translation $\mathbf{t}^{\text{h}}_i$, and shoulder translation $\mathbf{t}^{\text{s}}_i$ (each in $\mathbb{R}^{3}$), concatenated into $\mathbf{y}_i = [\mathbf{e}_i, \mathbf{r}^{\text{h}}_i, \mathbf{t}^{\text{h}}_i, \mathbf{t}^{\text{s}}_i] \in \mathbb{R}^{137}.$ Adopting this representation makes the comparison to \cite{Fallingwater} free of representation confound, since both predict the same signal and are rendered by the same decoder.

\textbf{System overview.} As shown in Figure~\ref{fig:system}, given a speech waveform, we synthesize a temporally coherent sequence of expression codes and rigid head pose at 25 Hz. Let $\mathbf{a} \in \mathbb{R}^{T \times 512}$ denote continuous audio embeddings from a frozen Mimi encoder~\cite{defossez2024moshi}, taken at 25 Hz frame grid of the target codes $\mathbf{y}_{1:T} \in \mathbb{R}^{T \times 137}$. The mapping is one-to-many, since many plausible facial performances accompany the same utterance: regression to the conditional mean yields near-stationary motion, so we train adversarially, using expression reconstruction only during warmup; head pose retains a $\ell_1$ term throughout training (Sec.~\ref{sec:training}). Variation is supplied by a noise latent $\mathbf{z}_{1:T} \in \mathbb{R}^{T \times 128}$, $\mathbf{z}_t \sim \mathcal{N}(\mathbf{0}, \mathbf{I})$. The generator $G(\mathbf{z}, \mathbf{a})$ is a transformer; two multi-scale convolutional discriminators, one audio-conditioned and one unconditional, supply the adversarial signal.

\subsection{Generator}
\label{sec:generator}
\textbf{Two-stack design}. $G$ separates what varies from what is determined by speech. A noise stack of $M{=}2$ layers processes $\mathbf{z}$ alone with bidirectional attention; a conditioning stack of $N{=}6$ causal layers then fuses the processed latent with audio. Causality bounds the input context only: predicted frames are never fed back, so the model is not autoregressive in its output and the full sequence is produced in a single forward pass. Placing the stochastic path before conditioning lets the latent develop temporally-correlated structure (a coherent motion style over a window) before speech constrains it, rather than perturbing an already-committed trajectory frame by frame. All layers are pre-norm blocks with RMSNorm~\cite{zhang2019rms}, rotary position embeddings~\cite{su2021roformer}, and SwiGLU feed-forward~\cite{shazeer2020glu}, with $d_{\text{model}}{=}512$, 8 heads, and $d_{\text{ff}}{=}4 d_{\text{model}}$.

\textbf{Audio conditioning}. Audio features are shifted left by $L$ frames so that output frame $t$ is conditioned on $\mathbf{a}_{t+L}$. A small lookahead ($L{=}1$) supplies the right-context for coarticulation, where a viseme anticipates the following phoneme, at a latency cost of 40 ms (see lookahead ablation in Sec.~\ref{sec:ablations}). Projected audio is added to the noise-stack output before the conditioning stack.

\textbf{Windowed attention}. Both stacks use windowed attention with radii $R_{\text{c}}$ (conditioning) and $R_{\text{n}}$ (noise); we set $R_{\text{c}}{=}R_{\text{n}}{=}64$. The conditioning stack attends causally over $[t{-}R_{\text{c}}{+}1, t]$ (2.6\,s), the noise stack symmetrically over $[t{-}R_{\text{n}}, t{+}R_{\text{n}}]$ (5.2\,s), making the receptive field independent of sequence length and admitting KV-cached incremental decoding. Stacking widens this: past dependence reaches $N(R_{\text{c}}{-}1) = 378$ frames of audio ($\approx$15\,s), while future dependence is only $M R_{\text{n}}$ frames of noise and $L$ of audio. Future noise costs no latency ($\mathbf{z}$ is sampled, not observed, so future noise is drawn ahead), leaving the $L$-frame audio lookahead as the only inference latency.

\textbf{Output head}. A single linear projection emits the $137$ channels of the motion representation. The pose rows of its bias are initialized to the dataset mean pose, so the model learns only the delta.

\subsection{Discriminators}
\label{sec:discriminators}
The \textbf{conditional} discriminator $D_c$ receives two channels, the code sequence and a linear projection of the audio to the same width, and so judges audio--motion correspondence; it applies the same shift as the generator, so each frame is paired with exactly the audio the generator saw. The \textbf{unconditional} discriminator $D_u$ receives the code sequence alone. When head pose is predicted it is appended to the expression channel, scaled per dimension by its dataset standard deviation to match the z-scored expression, and only $D_u$ sees this joint $137$-D signal; $D_c$ is restricted to the $128$-D expression.

Both discriminators view a clip as a 2-D image over (code $\times$ time) and are evaluated at temporal scales $S{=}\{1, 2, 4, 8\}$, via average-pooling along time, each judging a different temporal granularity, from per-frame articulation to multi-second head motion. Each scale has a sub-discriminator: five weight-normalized $3{\times}3$ convolutions with stride $(2,1)$, downsampling the feature axis while preserving temporal resolution, with LeakyReLU(0.2), plus a $1$-channel output convolution.

\subsection{Training objective}
\label{sec:training}
Both discriminators are trained with the multi-scale hinge loss over scales $s \in S$, and the generator with the corresponding term plus $\ell_1$ feature matching on the deepest convolutional activation of each scale; we write $\mathcal{L}^{\text{adv}}_c$, $\mathcal{L}^{\text{fm}}_c$ for the terms from $D_c$ and likewise for $D_u$. To prevent the generator ignoring $z$ we add the mode-seeking term of~\cite{mao2019mode}, $$\mathcal{L}_{\text{div}} = -\frac{d\big(e(z_1, a),\, e(z_2, a)\big)}{d(z_1,\, z_2) + \epsilon},$$ with $e(\cdot)$ the expression channels and $d$ the mean absolute difference; two latents are drawn per step. Head pose is supervised by an $\ell_1$ term $\mathcal{L}_{\text{hp}}$ and regularized by the squared first- and third-order time differences of the prediction, standardized per dimension ($\mathcal{L}_{\text{vel}}$, $\mathcal{L}_{\text{jit}}$). The full generator objective is
$$\mathcal{L}_G = \mathcal{L}^{\text{adv}}_c + \mathcal{L}^{\text{adv}}_u + \lambda_{\text{fm}}\big(\mathcal{L}^{\text{fm}}_c + \mathcal{L}^{\text{fm}}_u\big) + \lambda_{\text{div}}\mathcal{L}_{\text{div}} + \lambda_{\text{hp}}\mathcal{L}_{\text{hp}} + \lambda_{\text{vel}}\mathcal{L}_{\text{vel}} + \lambda_{\text{jit}}\mathcal{L}_{\text{jit}},$$
with $\lambda_{\text{fm}}{=}2$, $\lambda_{\text{div}}{=}1$, $\lambda_{\text{hp}}{=}1$, $\lambda_{\text{vel}}{=}0.01$, $\lambda_{\text{jit}}{=}0.05$. The first 5{,}000 steps use only the reconstruction ($\ell_1$ on expression and pose) and smoothness terms; generator and discriminator steps then alternate.

\section{Experiments}

\subsection{Experimental Settings}
\textbf{Dataset}. We use the dataset of~\cite{Fallingwater}: 34{,}906 clips from 832 subjects (469.6 hours) spanning speech and non-verbal vocalizations, partitioned into 784 training and 48 test identities. All results are reported on a fixed set of 64 held-out clips. 

\textbf{Implementation details}. We train generator and discriminators jointly for 200{,}000 steps on 8 NVIDIA H200 GPUs at batch size 128. Samples are 512-frame crops. The loss covers a 128-frame window whose offset is drawn uniformly from $[0,384]$ each step; frames before the window are generated to supply its causal history, so the model is supervised at every history depth up to the $378$-frame stacked reach. We use AdamW~\cite{loshchilov2019decoupled} at $1{\times}10^{-4}$ (generator) and $2{\times}10^{-4}$ (discriminators), cosine decay (factor $0.1$), and an EMA of the generator weights.

\subsection{Results}

We present quantitative and qualitative results. Unless stated otherwise, our model is evaluated in the final configuration of Sec.~\ref{sec:generator}: a single frame of audio lookahead ($L{=}1$, 40\,ms) and $R_{\text{c}}{=}R_{\text{n}}{=}64$. The lookahead ablation is reported in Sec.~\ref{sec:ablations}. We selected four state-of-the-art methods as baselines: Fallingwater \cite{Fallingwater}, DiffPoseTalk \cite{sun2024diffposetalk}, MemoryTalker \cite{MemoryTalker}, and ARTalk \cite{chu2024artalk}. All methods except MemoryTalker are stochastic, so for fair quantitative evaluation we draw 64 samples per sequence and average each metric over the draws. To assess distributional realism, we report Fr\'echet Expression Distance (FED) and Fr\'echet Pose Distance (FPD); lower is better. To assess expressiveness, we report within-clip variance ratios of the generated motion relative to ground truth; closer to 1.0 is better. We also report motion similarity to ground truth and lip-sync (Sync Score) following the Fallingwater evaluation protocol; higher is better. Table~\ref{tab:main-results} reports the quantitative results, and Figure~\ref{fig:phoneme} shows qualitative comparisons (additional examples can be found in Appendix~\ref{app:qualitative-comparisons}). All baselines are trained without style conditioning for direct comparability with our method; style-conditioned variants are reported in Appendix~\ref{app:style}.

\begin{figure}[h]
    \centering
    \includegraphics[width=0.95\textwidth]{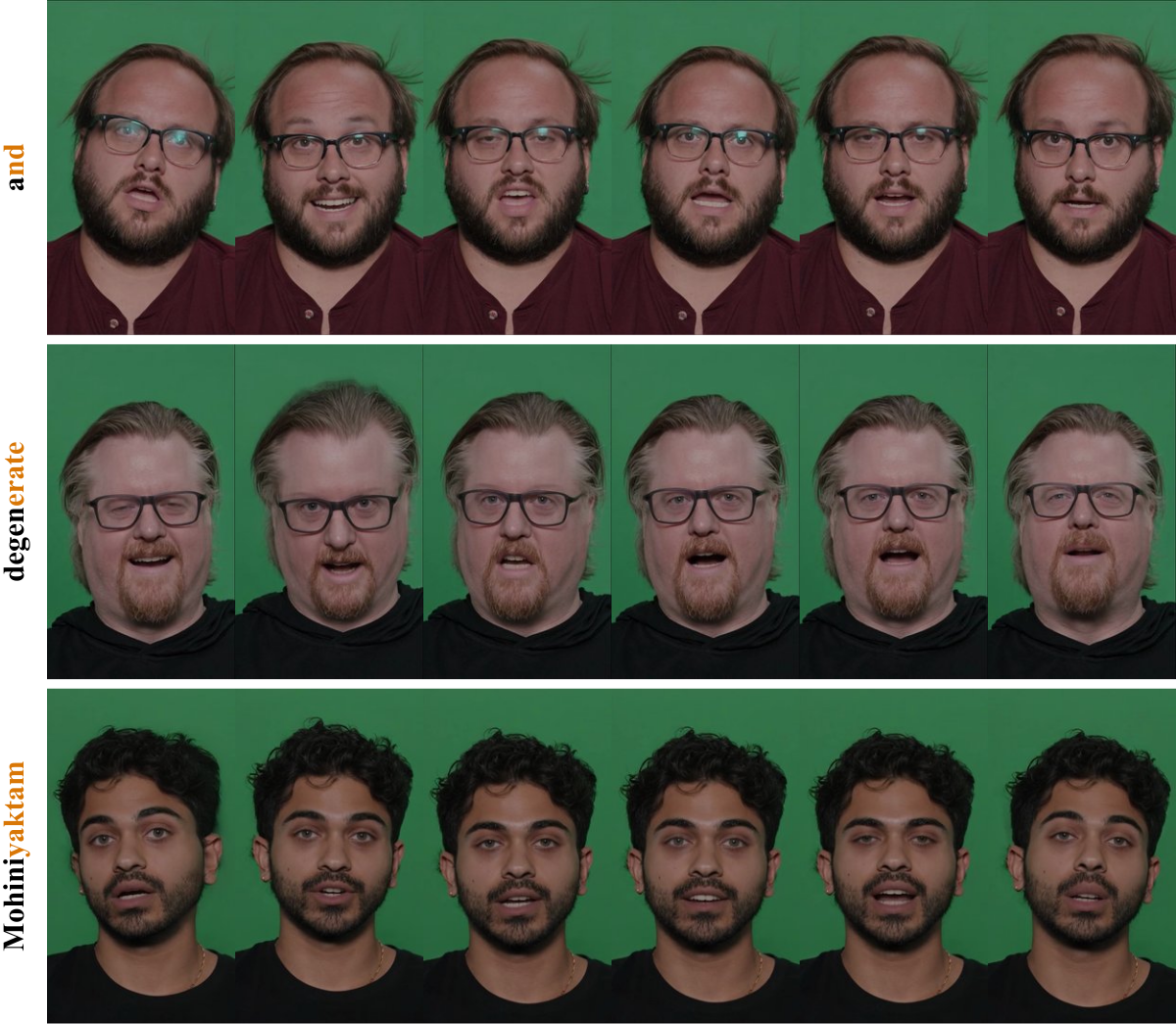}
    \\[6pt]
    {\footnotesize\hspace*{0.04\textwidth}\begin{tabular}{@{}c@{}c@{}c@{}c@{}c@{}c@{}}
    \makebox[0.16\textwidth][c]{\shortstack{Ground\\Truth}} & \makebox[0.16\textwidth][c]{\shortstack{Fallingwater\\\cite{Fallingwater}}} & \makebox[0.16\textwidth][c]{\shortstack{DiffPoseTalk\\\cite{sun2024diffposetalk}}} & \makebox[0.16\textwidth][c]{\shortstack{ARTalk\\\cite{chu2024artalk}}} & \makebox[0.16\textwidth][c]{\shortstack{MemoryTalker\\\cite{MemoryTalker}}} & \makebox[0.16\textwidth][c]{\shortstack{FaceGAN\\(ours)}}
    \end{tabular}}
    \caption{\label{fig:phoneme}Lip articulation on three utterances, with the target phoneme highlighted in each row label. Our mouth shapes track the ground-truth articulation faithfully with single-step inference, unlike the multi-step or non-causal baselines.}
    \vspace{-10pt}
\end{figure}

\begin{table}[t]
\centering
\small
\begin{tabular}{l|c|c|c|c|c|c}
Method & FED $\downarrow$ & FPD $\downarrow$ & Similarity $\uparrow$ & Sync Score $\uparrow$ & Expr Var $\to\!1$ & Pose Var $\to\!1$ \\\hline
Fallingwater  & \third{12.4824} & 1.6070 & 0.0976 & \second{0.6964} & \first{0.9803} & \third{1.1863} \\
DiffPoseTalk  & \first{11.3546} & \third{1.4720} & 0.1016 & \third{0.5768} & \third{0.7591} & 0.5354 \\
MemoryTalker  & 20.7047 & 3.7693 & \third{0.1328} & 0.5346 & 0.2171 & 0.0832 \\
ARTalk        & 14.5405 & \first{1.2727} & \first{0.2911} & 0.2342 & 1.3998 & \second{1.0539} \\
FaceGAN (ours) & \second{12.1483} & \second{1.3335} & \second{0.1542} & \first{0.8767} & \second{0.9647} & \first{0.9921}
\end{tabular}
\caption{\label{tab:main-results}Quantitative comparison against Fallingwater~\cite{Fallingwater}, DiffPoseTalk~\cite{sun2024diffposetalk}, MemoryTalker~\cite{MemoryTalker} and ARTalk~\cite{chu2024artalk}. Expr/Pose Var are within-clip variance ratios against ground truth (how much the face and head move over time during an utterance), so best is closest to $1.0$. \keyfirst = best, \keysecond = second-best, \keythird = third-best per column (ranking direction per arrow). Our method matches the multi-step or non-causal methods, achieving the best Sync Score and within-clip pose variance, while being the second best on every other metric, and operating in real time with single-step inference.}
\vspace{-10pt}
\end{table}

Our model substantially outperforms all baselines in Sync Score (Table~\ref{tab:main-results}) while being the only method that streams frame by frame at low latency. All other metrics are on par with the current state of the art, which means we did not trade speed for quality, unlike what usually happens with distilled diffusion models. This trade-off is a known artifact of diffusion distillation: DMD's reverse-KL objective behaves in a mode-seeking way, collapsing the generator onto dominant modes of the teacher distribution~\cite{yin2024dmd,li20261xdistillbreakingdiversityquality}.

\subsection{User study}
\label{sec:user-study}
To assess the perceptual quality of the generated face motion, we conducted a user preference study with 19 participants and a total of 684 comparisons. We use blind side-by-side A/B testing to compare our method against the baselines. We ask participants to focus on two criteria: lip sync (how well the mouth movements match the speech) and naturalness (whether the facial expressions and head motion look natural). We omit the identity-similarity criterion of~\cite{Fallingwater}, as our method does not condition on speaker style.

Pooled over all baselines, our method is at parity on lip sync and preferred on naturalness (Tab.~\ref{tab:user-study}). The per-opponent rows are reported for completeness, but with $76$ to $190$ comparisons each they resolve differences only above roughly ten points; the two that separate from chance, ARTalk on naturalness and MemoryTalker on both criteria, favour our method, and no baseline is ahead of ours on either criterion. The study agrees with the metrics: a single forward pass reaches the perceptual quality of the multi-step or non-causal baselines.

\begin{table}[h!]
\centering\small
\begin{tabular}{l|c|c}
Comparison & Lip sync & Naturalness \\\hline
vs.\ Fallingwater~\cite{Fallingwater}             & $45.3 \pm 7.1$ & $55.3 \pm 7.1$ \\
vs.\ DiffPoseTalk~\cite{sun2024diffposetalk}      & $52.1 \pm 7.1$ & $52.1 \pm 7.1$ \\
vs.\ ARTalk~\cite{chu2024artalk}                  & $45.6 \pm 9.1$ & $\mathbf{61.1 \pm 9.0}$ \\
vs.\ MemoryTalker~\cite{MemoryTalker}             & $\mathbf{67.1 \pm 10.6}$ & $\mathbf{68.4 \pm 10.5}$
\\\hline
all baselines                                     & $50.5 \pm 4.1$ & $\mathbf{57.1 \pm 4.1}$ \\
\end{tabular}
\caption{\label{tab:user-study}User study: percentage of comparisons preferring our method, with $95\%$ confidence intervals, over 19 participants and 684 comparisons. $50\%$ means the two are perceptually indistinguishable; bold marks intervals that exclude $50\%$.}
\vspace{-10pt}
\end{table}


\subsection{Ablation Study}
\label{sec:ablations}
We ablate the central design choices of our model: the temporal structure of the injected noise (Tab.~\ref{tab:noise}), and the audio lookahead, training objective and discriminator design (Tab.~\ref{tab:ablations}). All ablations are trained under identical settings; scores are on the 64-clip test set. Three independent training runs of the full model give the run-to-run band against which differences are judged. Alongside the distribution metrics we report sample diversity for expression and pose: with $K{=}64$ draws per clip, the spread across draws at a fixed frame, in units of the ground-truth temporal spread, so $1$ means two samples differ as much as a real face moves and $0$ means the model ignores $\mathbf{z}$.

\begin{table}[h]\centering\small\setlength{\tabcolsep}{4pt}
\begin{tabular}{ccc|rl|rl|rl|rl|rl}
\toprule
causal noise & radius & att. & \multicolumn{2}{c|}{Sync\,$\uparrow$} &
\multicolumn{2}{c|}{FED\,$\downarrow$} & \multicolumn{2}{c|}{Expr Div} & \multicolumn{2}{c|}{Pose Div} &
\multicolumn{2}{c}{Expr Var} \\
\midrule
\checkmark & 512 & 512 & 0.8654 & \rd{-2\%} & 11.59 & \gy{-1\%} & 0.89 & \gy{-3\%} & 0.84 & \gy{+4\%} &
0.92 & \rd{-8\%} \\
\checkmark & 129 & 129 & 0.8905 & \gy{+1\%} & 11.60 & \gy{-1\%} & 0.79 & \rd{-14\%} & 0.64 & \rd{-22\%} &
0.93 & \rd{-7\%} \\
\checkmark & 64 & 64 & 0.9074 & \gn{+3\%} & 12.61 & \gy{+7\%} & 0.68 & \rd{-25\%} & 0.52 & \rd{-36\%} &
0.92 & \rd{-9\%} \\
\midrule
& 2 & 5 & 0.8616 & \rd{-2\%} & 11.59 & \gy{-1\%} & 0.82 & \rd{-11\%} & 0.81 & \gy{-0\%} & 1.03 & \gy{+3\%}
\\
& 32 & 65 & 0.8902 & \gy{+1\%} & 11.64 & \gy{-1\%} & 0.91 & \gy{-0\%} & 0.91 & \gy{+12\%} & 1.02 &
\gy{+1\%} \\
ours & 64 & 129 & 0.8810 & {\scriptsize$\pm$0.0065} & 11.76 & {\scriptsize$\pm$0.52} & 0.91 &
{\scriptsize$\pm$0.03} & 0.82 & {\scriptsize$\pm$0.05} & 1.01 & {\scriptsize$\pm$0.02} \\
& 128 & 257 & 0.8847 & \gy{+0\%} & 11.88 & \gy{+1\%} & 0.91 & \gy{-0\%} & 0.88 & \gy{+8\%} & 1.02 &
\gy{+2\%} \\
\bottomrule
\end{tabular}
\caption{Noise stack, causal versus acausal; causal rows and ours are means of
three runs. ``att.'' is frames seen per layer: $R_{\text{n}}$ for a causal stack, $2R_{\text{n}}{+}1$ for an acausal one. Deltas against ours, whose $\pm$ is the run-to-run sd over its three runs; \gy{grey} is inside that band ($<2\sigma$). For Expr Div and Pose Div only collapse is marked, since higher diversity is not in itself better.}
\label{tab:noise}
\end{table}

\textbf{Noise shaping}. Sec.~\ref{sec:spectral_limitation} predicts that a causal generator driven by i.i.d.\ noise cannot place a spectral null without collapsing its innovation. Tab.~\ref{tab:noise} tests this by making the noise stack causal while holding everything else fixed. The two are not directly comparable at equal radius (a causal noise stack sees $R_{\text{n}}$ frames, an acausal one $2R_{\text{n}}{+}1$), so we compare at matched attended width and bracket our model with a narrower and a wider causal stack. All three causal stacks under-articulate, at $0.92$--$0.93$ against ground truth, while every acausal setting stays within $3\%$ of $1.0$. At matched width the causal stack also loses $14\%$ of expression diversity and $22\%$ of pose diversity at unchanged Sync and FED; the narrower stack loses $25\%$ and $36\%$.

\begin{table}[t]\centering\small\setlength{\tabcolsep}{3pt}
\begin{tabular}{ll|rl|rl|rl|rl|rl}
\toprule
& & \multicolumn{2}{c|}{Sync\,$\uparrow$} & \multicolumn{2}{c|}{FED\,$\downarrow$} & \multicolumn{2}{c|}{FPD\,$\downarrow$} & \multicolumn{2}{c|}{Expr Div} & \multicolumn{2}{c}{Pose Div} \\
\midrule
\multicolumn{2}{l|}{Ours ($L{=}1$)} & 0.8810 & {\scriptsize$\pm$0.0065} & 11.76 & {\scriptsize$\pm$0.52} & 1.62 & {\scriptsize$\pm$0.32} & 0.91 & {\scriptsize$\pm$0.03} & 0.82 & {\scriptsize$\pm$0.05} \\
\midrule
\multirow{3}{*}{Lookahead} & $L{=}0$ & 0.8698 & \gy{-1\%} & 13.08 & \rd{+11\%} & 2.38 & \rd{+46\%} & 1.03 & \gy{+13\%} & 0.99 & \gy{+21\%} \\
& $L{=}2$ & 0.8737 & \gy{-1\%} & 11.42 & \gy{-3\%} & 1.99 & \gy{+22\%} & 0.90 & \gy{-1\%} & 0.85 & \gy{+4\%} \\
& $L{=}4$ & 0.8721 & \gy{-1\%} & 11.82 & \gy{+1\%} & 2.24 & \gy{+38\%} & 0.90 & \gy{-2\%} & 0.81 & \gy{-0\%} \\
\midrule
\multirow{2}{*}{Discriminators} & no $D_u$ & 0.6403 & \rd{-27\%} & 11.98 & \gy{+2\%} & 4.44 & \rd{+173\%} & 1.21 & \gy{+33\%} & 0.08 & \rd{-90\%} \\
& no $D_c$ & 0.0234 & \rd{-97\%} & 15.16 & \rd{+29\%} & 1.87 & \gy{+15\%} & 1.21 & \gy{+33\%} & 0.86 & \gy{+5\%} \\
\midrule
\multirow{3}{*}{Objective} & $\lambda_{\text{fm}}{=}0$ & 0.8602 & \rd{-2\%} & 16.48 & \rd{+40\%} & 1.94 & \gy{+20\%} & 1.30 & \gy{+42\%} & 1.17 & \gy{+43\%} \\
& $\lambda_{\text{div}}{=}0$ & 0.8851 & \gy{+0\%} & 11.97 & \gy{+2\%} & 2.79 & \rd{+72\%} & 0.09 & \rd{-91\%} & 0.07 & \rd{-91\%} \\
& $\lambda_{\text{hp}}{=}0$ & 0.8625 & \rd{-2\%} & 11.18 & \gy{-5\%} & 1.58 & \gy{-3\%} & 0.97 & \gy{+7\%} & 1.00 & \gy{+22\%} \\
\bottomrule
\end{tabular}
\caption{Ablations on the 64-clip test set. Deltas are percentages against ours (mean of three runs), whose $\pm$ is the run-to-run sd over three runs; \gy{grey} marks differences inside that band ($<2\sigma$). For Expr Div and Pose Div only collapse is marked, since higher diversity is not in itself better.}
\label{tab:ablations}
\vspace{-10pt}
\end{table}

\textbf{Lookahead, objective and discriminators}. Lookahead beyond one frame changes nothing, while $L{=}0$ costs expression and pose quality. The architecture therefore supports fully causal operation ($L{=}0$) where required, trading expression and pose quality for 40\,ms. Both discriminators are necessary: without $D_c$ lip-sync collapses entirely ($-97\%$ Sync) while pose is untouched, and without $D_u$ pose collapses in both distribution and diversity ($+173\%$ FPD, $-90\%$ Pose Div) and lip-sync degrades as well. Removing feature matching inflates FED by $40\%$. The mode-seeking term has a measurable effect on the metric it targets: removing it collapses diversity by $91\%$ in both expression and pose: the generator reproduces a single performance regardless of $\mathbf{z}$, and FPD inflates by $72\%$. Removing the pose $\ell_1$ term is tolerable, costing $2\%$ Sync.

\subsection{Computational performance}
We report the computational performance for generator alone, excluding frozen audio encoder and renderer. All numbers are measured in streaming mode (one frame per call, with each layer keeping a fixed-size ring buffer over its attention window, so per-frame cost and memory are independent of stream length) on a single H200 in fp32, with no optimisation beyond the KV cache. Latency is input-to-output: the $L{=}1$ lookahead ($40$\,ms) plus the forward pass. A single stream runs at $43.3$\,ms, $12.1\times$ real time. Batching is nearly free: latency is unchanged at $256$ concurrent streams, still under $50$\,ms at $1024$, and real time holds to $4096$ streams at $62$\,ms (Tab.~\ref{tab:perf}). Coarser emission trades latency for throughput: at $1$\,s blocks the per-call cost amortises to $235\times$ real time, at $1$\,s latency.
  
\begin{table}[h]
\centering\small
\begin{tabular}{l|rrrrr}
\toprule
concurrent streams & 1 & 256 & 1024 & 2048 & 4096 \\
\midrule
forward / latency (ms) & 3.3 / 43.3 & 3.6 / 43.6 & 6.5 / 46.4 & 11.4 / 51.4 & 21.7 / 61.7 \\
$\times$\,real time & 12.1 & 11.0 & 6.2 & 3.5 & 1.8 \\
\bottomrule
\end{tabular}
\caption{Latency and throughput of the generator on one H200. Latency is the
forward pass plus the $40$\,ms audio lookahead; real-time factor is against the
$40$\,ms frame budget at 25\,Hz.}
\label{tab:perf}
\vspace{-10pt}
\end{table}

\section{Conclusion}
We showed that audio-driven facial animation does not need iterative sampling. On this task a single-pass GAN matches the compared baselines. Because generation is one forward pass over a causal audio stack with a single frame of lookahead, the same weights run offline and streaming, at $43$\,ms latency for a single stream and $62$\,ms at $4096$ concurrent streams on one GPU. The ablations locate where the quality comes from: both discriminators are necessary and in different ways, and the noise stack must be acausal: a causal one under-articulates and loses sample diversity, which is free to avoid, since noise is sampled rather than observed.

\newpage

\subsection*{AI use statement}
We used AI tools to aid and polish writing, e.g. to save space by compacting long paragraphs, enhance the clarity of sentences, format tables, etc. 
We also used AI tools to draft sections of the paper, e.g. we used model config to draft implementation details, and result data to create LaTeX tables. 
We have reviewed all AI-assisted work, e.g. we proof-read AI modified paragraphs, checked that reported numbers match the experiment results, etc. 
All conceptual contributions, like use of GANs and causal noise shaping, the experimental designs, ablation and analyses are the sole work of the authors.
We take responsibility for the final content of this work, including text, claims or artifacts produced with the aid of generative AI.

\subsection*{Ethics statement}
Generation of photorealistic talking heads raises concerns about the potential for misuse, like deepfake creation. 
Two properties of our system bound that risk without removing it: the model outputs low-dimensional expression and head-pose coefficients rather than pixels, so photorealistic video requires renderer that we do not contribute; and it is not conditioned on speaker identity or style. On the other hand, real-time single-pass operation lowers the cost of misuse relative to existing methods. We hope to raise awareness of risks and encourage productive discussions regarding fair uses of synthetic video generation models.

\bibliographystyle{iclr2027_conference}
\bibliography{sample}


\clearpage
\appendix
\section{Background for Spectral Limitation}
\label{app:prereq}

\paragraph{Stationary processes and spectra.}
$\{y_t\}_{t\in\mathbb{Z}}$ is $L^2$-stationary if $\mathbb{E}[y_t^2]<\infty$, $\mathbb{E}[y_t]=\textit{const.}$, and $\mathrm{Cov}(y_t,y_{t+h})=\gamma(h)$ depends only on the lag $h$ (i.e.\ the process has finite variance and its first and second moments do not drift with absolute time). In particular the process looks statistically the same at every $t$, which holds here
because the generator \eqref{eq:generator} is time-invariant and driven by i.i.d.\ noise. Its spectral density $S\geq 0$ distributes variance over frequency: $\mathrm{Var}(y_t)=\frac{1}{2\pi}\int_{-\pi}^{\pi}S(\omega)\,d\omega$. A perfect stopband is $S=0$ on a set of positive Lebesgue measure (e.g.\ a frequency interval). The main text uses $S$ and this stopband definition to state Szeg\H{o}'s formula and its key consequence.

\paragraph{Predictability.}
$\sigma_\infty^2$ is the smallest possible mean-squared error predicting $y_t$ from
linear combinations of past values $y_{<t}=(y_{t-1},y_{t-2},\dots)$. It measures how much genuinely new randomness the process generates per step. If $\sigma_\infty^2=0$, $y_t$ is perfectly recoverable from its past (i.e.\ it is deterministic), so $\mathrm{Var}(y_t\mid y_{<t})=0$.  An element of the closed linear span ($L^2$-limits of past linear combinations) is past-measurable, hence a function of $y_{<t}$ a.s. In the main text this is used to turn $\sigma_\infty^2=0$ into $\mathrm{Var}=0$, which the main proof then contradicts by showing $\mathrm{Var}(y_t\mid y_{<t})>0$ a.s. 

\paragraph{Fresh-noise usage.}
The generator genuinely uses its current input if holding past noise fixed and varying $z_t$ still varies the output,
\begin{equation}
  \mathrm{Var}(F(z_t,z_{<t})\mid z_{<t})>0 \quad \text{a.s.}
  \label{eq:nondeg}
\end{equation}
For example, $y_t=z_{t-1}$ would violate \eqref{eq:nondeg}, while $y_t=z_t+0.5z_{t-1}$ satisfies it. Any non-trivial generator sampling fresh noise per frame satisfies it. Used as the standing assumption of the main
claim; the main proof cites it to force $\mathrm{Var}(y_t\mid z_{<t})>0$.

\paragraph{Nested information.}
Each $y_s$, $s<t$, is computed from $z_{<t}$ alone, so past outputs are deterministic functions of past
noise: knowing $z_{<t}$ tells you $y_{<t}$, but not vice versa (the map is many-to-one for nonlinear $F$). In information terms $\sigma(y_{<t})\subseteq\sigma(z_{<t})$, i.e.\ $\sigma(y_{<t})$ is the less informative of the two. The law of total variance for $\mathcal{G}\subseteq\mathcal{F}$ gives
\begin{equation}
\mathrm{Var}(y_t\mid y_{<t})=\mathbb{E}[\mathrm{Var}(y_t\mid z_{<t})\mid y_{<t}]
+\mathrm{Var}(\mathbb{E}[y_t\mid z_{<t}]\mid y_{<t})
\geq \mathbb{E}[\mathrm{Var}(y_t\mid z_{<t})\mid y_{<t}].
\label{eq:totalvar}
\end{equation}
Both terms are non-negative, so dropping the second yields the lower bound, used in the first line of the main proof to lift variance given noise-past to variance given output-past.

\section{Style-Conditioned Results}
\label{app:style}

The main paper compares all baselines without style conditioning for direct
comparability with our method. This appendix reports the same metrics for
the style-conditioned variants, plus our (unconditioned) model as a reference. The results are shown in Table~\ref{tab:style-quality}.

\begin{table}[h]
\centering
\small
\begin{tabular}{l|c|c|c|c|c|c|c}
Method & style & FED $\downarrow$ & FPD $\downarrow$ & Similarity $\uparrow$ & Sync Score $\uparrow$ & Expr Var& Pose Var \\\hline
Fallingwater  & \checkmark & \third{12.3718} & 1.7186 & \first{0.7052} & \second{0.6977} & \second{1.1058} & \third{1.2905} \\
DiffPoseTalk  & \checkmark & \first{9.8356} & \first{1.3321} & \second{0.5683} & 0.5100 & \third{0.6428} & 0.4661 \\
MemoryTalker  & \checkmark & 16.2776 & 3.9006 & 0.1486 & \third{0.5722} & 0.2651 & 0.0227 \\
ARTalk        & \checkmark & 12.4887 & \third{1.3644} & \third{0.5223} & 0.3121 & 1.4379 & \second{0.8492} \\
FaceGAN (ours) & & \second{12.1483} & \second{1.3335} & 0.1542 & \first{0.8767} & \first{0.9647} & \first{0.9921}
\end{tabular}
\caption{\label{tab:style-quality}Style-conditioned comparison against Fallingwater~\cite{Fallingwater}, DiffPoseTalk~\cite{sun2024diffposetalk}, MemoryTalker~\cite{MemoryTalker} and ARTalk~\cite{chu2024artalk}; \checkmark marks style conditioning. Expr/Pose Var are within-clip variance ratios against ground truth (how much the face and head move over time during an utterance), so best is closest to $1.0$. \keyfirst = best, \keysecond = second-best,
\keythird = third-best per column (ranking direction per arrow).}
\end{table}

\section{Additional Qualitative Comparisons}
\label{app:qualitative-comparisons}
\begin{figure}[h]
    \centering
    \includegraphics[width=0.95\textwidth]{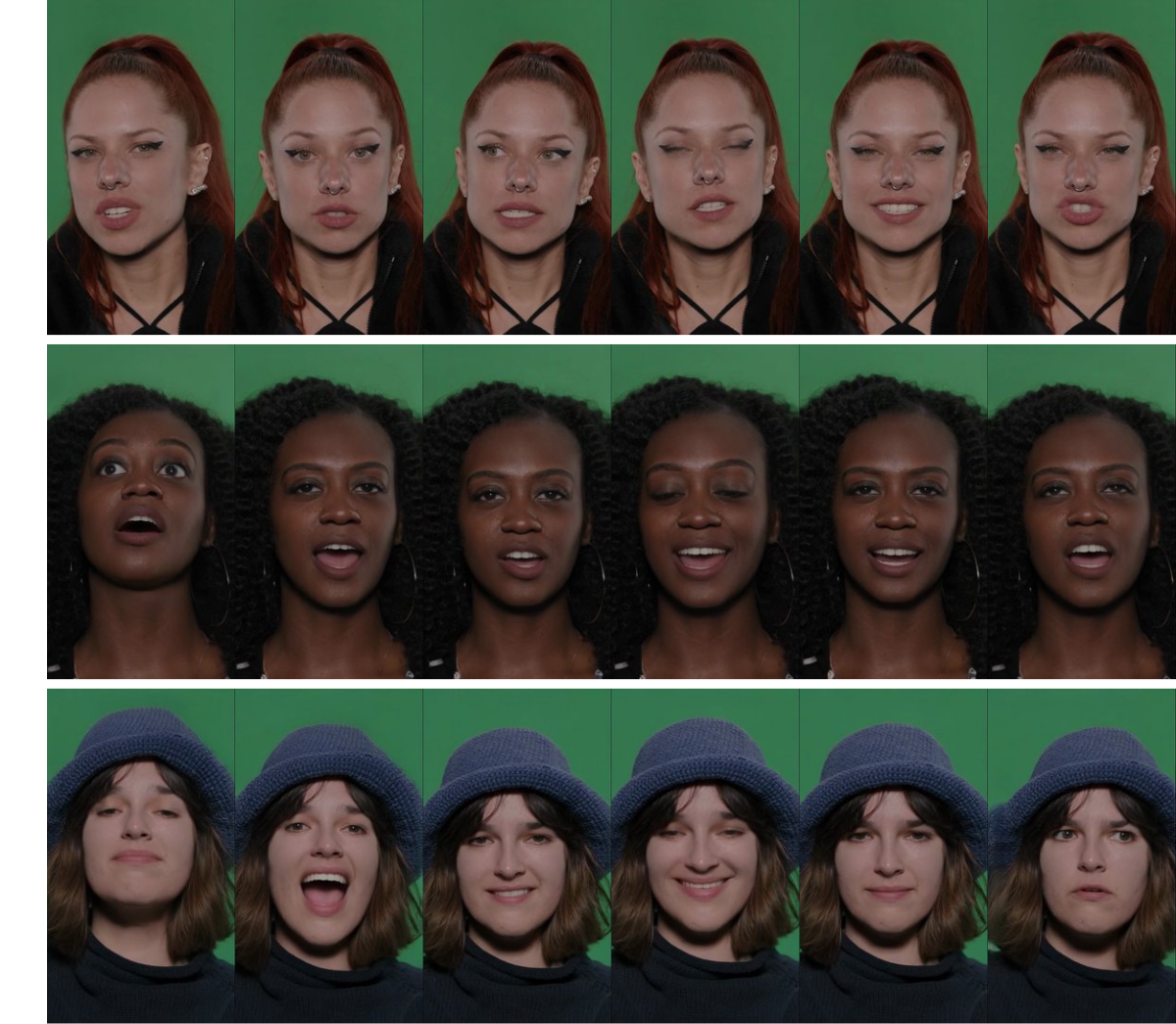}
    \\[6pt]
    {\footnotesize\hspace*{0.04\textwidth}\begin{tabular}{@{}c@{}c@{}c@{}c@{}c@{}c@{}}
    \makebox[0.16\textwidth][c]{\shortstack{Ground\\Truth}} & \makebox[0.16\textwidth][c]{\shortstack{Fallingwater\\\cite{Fallingwater}}} & \makebox[0.16\textwidth][c]{\shortstack{DiffPoseTalk\\\cite{sun2024diffposetalk}}} & \makebox[0.16\textwidth][c]{\shortstack{ARTalk\\\cite{chu2024artalk}}} & \makebox[0.16\textwidth][c]{\shortstack{MemoryTalker\\\cite{MemoryTalker}}} & \makebox[0.16\textwidth][c]{\shortstack{FaceGAN\\(ours)}}
    \end{tabular}}
    \caption{\label{fig:headpose}Head pose variation across three identities.}
\end{figure}

As a part of the supplemental material we include videos that show comparison with the baselines. In addition, Figure~\ref{fig:headpose} shows another qualitative single-frame example. 

\section{Inference Characteristics in Detail}
\label{app:inference-steps}

Table~\ref{tab:inference} in the main paper summarizes the inference
characteristics of the compared methods. Table~\ref{tab:inference-detail}
below gives the per-method verification behind the \#-steps counts.

\begin{table*}[h]
  \centering
  \footnotesize
  \setlength{\tabcolsep}{3pt}
  \begin{tabular}{l|l|c|c|c|>{\raggedright\arraybackslash}p{0.34\textwidth}}
  Method & Arch. & \# steps & Causal & \shortstack{Live-system\\ready} & Step counting \\\hline
  DiffPoseTalk & Diffusion & 500 & \textcolor{orange}{$\sim$} (chunk-level) & \textcolor{red}{\ding{55}} & Full 500-level DDPM ancestral schedule per 100-frame chunk; one denoising pass per level, no step-skipping \\
  ARTalk & LLM & 5 & \textcolor{orange}{$\sim$} (chunk-level) & \textcolor{red}{\ding{55}} & Coarse-to-fine AR schedule over 5 token scales ([1, 5, 25, 50, 100]); one pass per scale \\
  MemoryTalker & Transformer & 1 & \textcolor{red}{\ding{55}} & \textcolor{red}{\ding{55}} & Single deterministic forward pass over the full sequence \\
  Fallingwater & LLM & 176 & \textcolor{orange}{$\sim$} (chunk-level) & \textcolor{red}{\ding{55}} & Per-token AR over 176 tokens across 4 scales ([1, 25, 50, 100]), with style-bank cross-attention per step \\
  \textbf{FaceGAN (ours)} & \textbf{GAN} & \textbf{1} & \textbf{\textcolor{green!55!black}{\ding{51}}} (frame-level) & \textbf{\textcolor{green!55!black}{\ding{51}}} & One generator pass per frame; causal attention with 40\,ms lookahead
  \end{tabular}
  \caption{\label{tab:inference-detail}How \# steps is counted. For each method, \# steps is the number of sequential forward evaluations of the core generative network needed to finalize one generation unit, independent of clip length. DiffPoseTalk runs the full 500-level DDPM ancestral schedule per 100-frame chunk: one denoising pass per level, with no DDIM or step-skipping. ARTalk generates each 100-frame chunk through a coarse-to-fine autoregressive schedule over 5 token scales (patch sizes [1, 5, 25, 50, 100]), invoking the network once per scale. Fallingwater decodes its 4-scale pyramid (patch sizes [1, 25, 50, 100]) token by token, invoking the network 176 times per chunk with style-bank cross-attention at each step. MemoryTalker produces the full sequence in a single deterministic forward pass (audio encoder $\to$ memory retrieval $\to$ decoder). FaceGAN emits one frame per forward pass of the generator.}
\end{table*}

\section{Real-Time Benchmark of the Compared Methods}
\label{app:realtime-benchmark}

Table~\ref{tab:realtime-cost} reports generation compute and latency for every compared method on an NVIDIA H200 (torch 2.12.1, fp32, batch size 1, 30\,s of speech), excluding model loading, audio encoding, rendering, and I/O. Measurements were reproduced on two independent H200 nodes agreeing within 2.2\%. FaceGAN was timed without loaded weights; its decode has no data-dependent branching, so timing is weight-independent.

\begin{table}[t]
\centering\small
\begin{tabular}{llrr}
\toprule
Method & Decoding & Compute & Latency $\downarrow$ \\
\midrule
MemoryTalker & one-shot, non-causal           & 0.02\,s & full utterance \\
DiffPoseTalk & 4\,s chunk, 500-step diffusion  & 11.2\,s & 5.56\,s \\
Fallingwater & 4\,s chunk, 176-step AR         &  9.0\,s & 5.29\,s \\
ARTalk       & 4\,s chunk, 5-step AR           &  0.3\,s & 4.20\,s \\
\midrule
\textbf{Ours} & \textbf{per-frame}             & \textbf{2.4\,s} & \textbf{83\,ms} \\
\bottomrule
\end{tabular}
\caption{\label{tab:realtime-cost}Compute is total wall-clock time to generate 30\,s of motion (motion generation only, H200, fp32, batch 1); values below 30\,s are faster than real time. Latency is the delay before the first frame can be emitted: for the chunked methods, 4.00\,s of audio buffering $+$ per-chunk compute $+$ 0.16\,s of centred Savitzky--Golay look-ahead; for ours, one 40\,ms frame $+$ 40\,ms acausal look-ahead $+$ 3.2\,ms compute.}
\end{table}

Every method clears 25 fps; what differs is latency. ARTalk reports an RTF of 0.01 and describes this as low latency~\cite{chu2024artalk}; we reproduce that throughput (RTF 0.010). Throughput, however, is not latency: its 100-frame window means no frame can be emitted until 4\,s of audio has arrived, giving a time-to-first-frame of 4.20\,s. Ours emits the first frame in 83\,ms, about 50$\times$ less; at the same 4\,s latency it is 9$\times$ faster than ARTalk (880$\times$ real time) with no retraining. None of the baselines is therefore deployable for live animation. Lookahead horizons differ by two orders of magnitude: 40\,ms for ours, up to 4\,s for the chunked baselines, unbounded for MemoryTalker; block size (1--250) is a runtime flag in ours.

The 4\,s floor is architectural: ARTalk and Fallingwater decode each 100-frame window into a coarse-to-fine token pyramid (ARTalk~\cite{chu2024artalk}: [1, 5, 25, 50, 100]; Fallingwater: [1, 25, 50, 100]) in scale order rather than time order, with the coarsest scale a single token spanning the whole window, while DiffPoseTalk runs 500 diffusion steps per fixed 100-frame window. In ARTalk the window is welded to learned weights, so shrinking it requires retraining; in Fallingwater it is a config choice with length-agnostic encoding, so a smaller window would run with untested quality. Both implementations also zero-pad short clips to a chunk multiple, with untested quality effects.

Fallingwater is about 31$\times$ slower than ARTalk despite fewer parameters (71.8M versus 489.5M): 5 scale-parallel steps versus 176 sequential token steps per chunk. Lowering classifier-free guidance from 2.0 to 1.0 changes throughput by about 1\%, so sequential depth, not guidance, is the bottleneck.

\end{document}